\documentclass{article} 
\usepackage{iclr2018_conference,times}
\usepackage{hyperref}
\usepackage{url}
\usepackage{dsfont}
\usepackage{amsmath,amssymb,amsthm}
\usepackage{pgffor}
\usepackage{enumerate}
\usepackage{booktabs}
\usepackage{subcaption}
\usepackage{placeins}

\usepackage{todonotes}

\title{The best defense is a good offense: \\ Countering black box attacks by predicting slightly wrong labels}

\author{Yannic Kilcher \\
Department of Computer Science\\
ETH Zurich\\
\texttt{yannic.kilcher@inf.ethz.ch}
\And
Thomas Hofmann \\
Department of Computer Science\\
ETH Zurich\\
\texttt{thomas.hofmann@inf.ethz.ch}
}

\iclrfinalcopy 
\arxiv

\begin{document}

\maketitle

\begin{abstract}
    Black-Box attacks on machine learning models occur when an attacker, despite having no access to the inner workings of a model, can successfully craft an attack by means of model theft.
    The attacker will train an own substitute model that mimics the model to be attacked.
    The substitute can then be used to design attacks against the original model, for example by means of adversarial samples.
    We put ourselves in the shoes of the defender and present a method that can successfully avoid model theft by mounting a counter-attack.
    Specifically, to any incoming query, we slightly perturb our output label distribution in a way that makes substitute training infeasible.
    We demonstrate that the perturbation does not affect the ordinary use of our model, but results in an effective defense against attacks based on model theft.
\end{abstract}

\section{Introduction}

Think of a situation where a company has invested a lot of resources into training a proprietary machine learning model and are now giving their customers access to query the model via a paid API\@.
In this case, the company would like to protect its trained model as intellectual property, both because of the need to recoup the sunk cost of building it and also because releasing the model in full could expose privacy-relevant information about the training data.
However, recent work has shown that it is possible for an attacker to extract enough information about the proprietary model by simply observing its inputs and outputs, such that the attacker can essentialy forge an identical copy of that model.
This procedure is called \emph{model theft}, where an attacker, without access to some model, can copy the model by observing input and output to it and then train an own model to mimic the inaccessible model's behavior.

We focus on the work of~\cite{papernot2016practical} that crafts adversarial samples against a black-box classifier.
An adversarial sample~\citep{szegedy2013intriguing} is an input sample to a classifier model that has been perturbed in a manner imperceptible to a human, but fools the classifier into assigning a wrong output label with very high probability.
In this context, \emph{black-box} means that the attacker does not have access to the internals of the model to be attacked, but can only observe its outputs to given inputs.
Such internals, especially gradient computations, are required by modern methods to craft aversarial samples.
However, by using the property of \emph{transferability}, meaning adversarial samples crafted for some model trained on some data are also likely to be adversarial samples for a second model trained on the same data, the authors can successfully attack the classifier without the needed internal access.
They do this by training a substitute model, for which they do have internal access, and then use this substitute to craft adversarial samples for the model to be attacked.
This one particular example of model theft, with the end goal to use the copied substitute model to generate adversarial samples for the black-box.

More than simply observing input and output,~\cite{papernot2016practical} introduce a method called \emph{jacobian data augmentation} to speed up the process of model theft.
Much like the construction of adversarial samples, the authors carefully craft input samples to the black-box model such that they will gain maximal information about the model's decision boundaries from the label response.
This allows them, after starting with just a tiny labeled seed set (150 examples), to successfully attack the black-box model after a relatively short time.
The authors discuss several strategies to counter their attack, focusing mainly on methods known to protect against adversarial samples, but do not find any of them to be successful in fully averting their attack.

In this work, we present a method to reliably counter such black-box attacks based on model theft.
As such, throughout the discussion, we shall view the world from the eyes of the defender.
Thus \emph{our} model is the black-box to be attacked and the \emph{attacker} is training a substitute, possibly using jacobian data augmentation to do so.

Where as all previously proposed countermeasures focus on meddling with the inputs to our model, we focus on changing its outputs.
Since the attacker is actively using our outputs in training the substitute, this channel of information provides us with the opportunity to launch a counter-attack by predicting wrong labels to throw off the substitute training.
In a sense, we perform a variant of \emph{training set poisoning} for the attacker, but rather than poisoning the input data, we poison the labels.

But does this mean that we effectively sacrifice part of our API's classification accuracy?
After all, our non-malicious customers are paying for the most accurate classification we can offer and degrading our output labels would be of great disservice to them.
To avoid this, we introduce the only deviation of our setup to the one considered by~\cite{papernot2016practical}.
Where as the previous authors attack black-box models where only the binary output label is observable, we focus on the setting where the attacker can observe the output label distribution.
We do this for two reasons:
First, this is generally how modern classifier APIs operate, giving not one, but the top $k$ output labels, each one with some percentage score.
Second, this allows us to counter a model thief by changing our output label distribution such that the effective class prediction (the argmax) remains unchanged while the distribution itself changes only minimally and any non-malicious consumer remains unaffected.\footnote{Naturally, an effective defense against our counter-attack is therefore for a model thief to simply re-binarize the output label distribution and revert to the original black-box attack method. While this is certainly true, consider that an attacker can train a substitute much more effectively when matching the output label distribution than when simply matching the binarized labels, so defending against our counter-attack, though possible, can be very costly.}

So how do we change our output label distribution to prevent black-box attacks?
For this, consider the fact that to the defender, the attacker's substitute model is also a black-box that can be attacked using the exact same method:
Train a substitute model for the attacker's substitute model, call it a counter-substitute, and use it to craft adversarial examples for the attacker.
However, there are several problems with this approach.
First, we need to counter the attacker's substitute while it is still in training, meaning that the model itself is non-static and we have no way of knowing how far along in training it is.
Second, we would like to counter any incoming attack from any variety of model, so it is not enough to simply have a single substitute model as we need to consider entire families of them.
As a consequence, it looks like our only option is to maintain a huge zoo of substitute models, using different architectures and training each one from multiple initializations for varied amounts of progress up to convergence, remember all these models and then try to counter-attack all of them.

Here is where we have a trick up our own sleeve.
Consider the following question:
What is the attacker's single objective?
To copy our model.
Therefore, what is the single best counter-substitute to any attacker's substitute model?
Our model that they are trying to copy.
By crafting adversarial samples \emph{to ourselves}, we can reliably throw off any attacker attempting model theft without having to do any of the additional work of training approrpiate counter-substitutes.
Our method is deceptively simple, can be plugged into any existing classifier and delivers effective protection against model theft with minimal overhead and without affecting its ordinary use.
 
\section{Method}

\subsection{Adversarial example crafting}

In this work, we focus mainly on the \emph{Fast Gradient Sign Method} proposed in~\cite{goodfellow2014explaining}, both because it is widely used and because~\cite{papernot2016practical} found it to be very effective amongst the methods they explored for performing black-box attacks.
Given a model $F$, a cost function $c$, and an input-output pair $x, y$, the Fast Gradient Sign Method constructs an adversarial example for $x$ as
\begin{equation}
    x^* = x + \epsilon~\text{sign}(\nabla_x c(F, x, y))
\end{equation}

This maximizes the cost function $c$ while constraining the adversarial perturbation to $\|\bullet\|_\infty \leq \epsilon$.
Note that the Fast Gradient Sign Method can be computed easily using a single pass of backpropagation and therefore creates minimal overhead.

\subsection{Generating the counter-attack}

In order to effectively counter a model thief, we must first define what exactly this means.
What we want is to counter the thief while they are training their substitute model.
They do this training by using some gradient descent method on the weights of their substitute model given the input-output pairs from our model.
Therefore, to throw off the thief's training procedure, we should focus on maximally messing up their training gradients.
Since gradients tend to get smaller in magnitude as one approaches a cost function's optimum, our goal should be to perturb our output label distribution in such a way that an attacker's training gradients will be of very large magnitude.

Formally, given a model $F_\theta$ with parameters $\theta$, a cost function $c$, a given input $x$ and our (true) predicted label distribution $y_x$, our adversarial counter-attack generation process is as follows:
\begin{equation}
    y_x^* = y_x + \epsilon~\text{sign}(\nabla_y \|\nabla_\theta c(F_\theta, x, y_x)\|_2)
\end{equation}

Of course, in our case, the model $F_\theta$ would be our own classifier, since, as explained above, it is the best counter-substitute for any model thief's substitute model.
By varying $\epsilon$, we have a direct handle on how much of an impact this corruption will have on the quality of our final output.
We have found values as low as $\epsilon = 0.003$ are already enough to mount a successful counter-attack.\footnote{Note that, in general, this procedure will not yield a normalized output distribution, which is only partly relevant, since common APIs classifying among many different class labels usually don't output the full label distribution, but only a handful of top scoring labels.
In case normalization is important, we describe a number of normalization methods that we have found to not interfere with the counter-attack in the Appendix.
One can also easily come up with advanced methods for creating adversarial samples that take the normalization and positivity requirements of the output into account when crafting the samples.
}
 
\FloatBarrier
\section{Experiments}

\subsection{Setup}
Our experimental setup is akin to~\cite{papernot2016practical}, except that we don't attack an actual public API, since we're now in the shoes of the defender.
Thus, our defending model is a simple 3-layer CNN and we attack it using a variety of substitute models, all using somewhat different architectures.
Table~\ref{tbl:models} shows the model architectures used.
This, again, is done akin to~\cite{papernot2016practical}, where we switched out few architectures because we found some of their proposed architectures to not work in our setting.

We train our defending model on MNIST, achieving an accuracy of roughly 99\%.
Results on further datasets are given in the Appendix.
As in the original work, the attacker is given 150 labeled samples of the true data in order to perform bootstrapping and has to perform dataset augmentation to obtain more samples.
We perform 6 rounds of data augmentation, each one doubling the substitute model's available dataset in size.

We use the CleverHans library~\citep{papernot2016cleverhans} to generate all adversarial samples.
For each experimental setting we report means and one standard deviation over 10 repetitions.

\begin{table}[ht!]
    \centering
    \begin{tabular}{cccccccc}
        \toprule
        ID & Conv & Conv & Conv & FC & FC & FC & S \\
        \midrule
        X & 64 & 128 & - & - & - & - & 10 \\
        Y & 64 & 128 & 128 & - & - & - & 10 \\
        A & 32 & 64 & - & 200 & 200 & - & 10 \\
        F & 32 & 64 & - & 200 & - & - & 10 \\
        G & 32 & 64 & - & - & - & - & 10 \\
        H & 32 & - & - & 200 & 200 & - & 10 \\
        I & - & - & - & 200 & 200 & 200 & 10 \\
        J & - & - & - & 1000 & 200 & - & 10 \\
        K & - & - & - & 1000 & 500 & 200 & 10 \\
        L & 32 & - & - & 1000 & 200 & - & 10 \\
        \bottomrule
    \end{tabular}
    \caption{Substitute model architectures used for the attacker. \emph{Conv}: Convolutional layer. \emph{FC}: Fully connected layer. \emph{S}: Softmax layer.}
    \label{tbl:models}
\end{table}

\FloatBarrier
\subsection{Results}
\begin{figure}[ht!]
    \centering
    \begin{subfigure}[t]{0.49\textwidth}
        \centering
        \includegraphics[width=\textwidth]{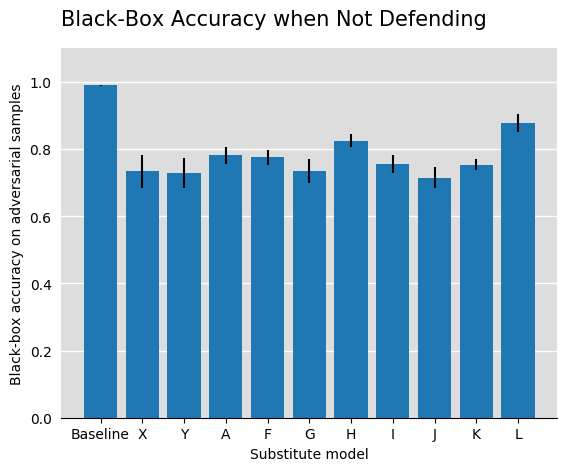}
    \end{subfigure}
    \begin{subfigure}[t]{0.49\textwidth}
        \centering
        \includegraphics[width=\textwidth]{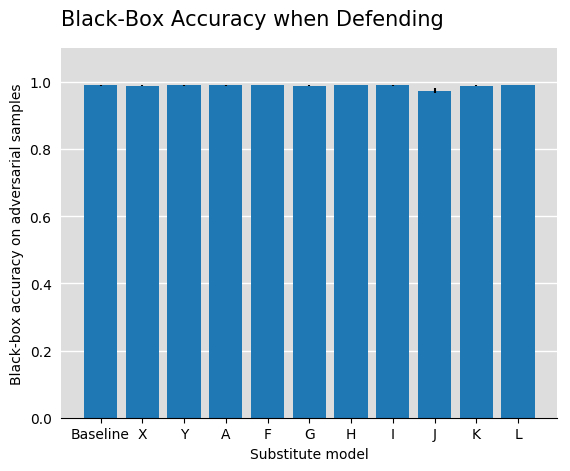}
    \end{subfigure}
    \par\bigskip
    \begin{subfigure}[b]{0.49\textwidth}
        \centering
        \includegraphics[width=\textwidth]{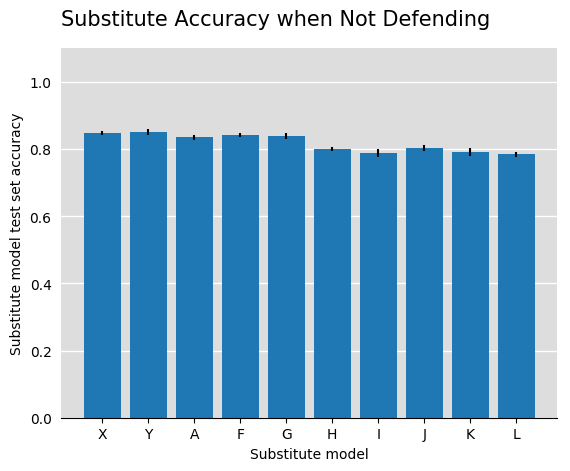}
    \end{subfigure}
    \begin{subfigure}[b]{0.49\textwidth}
        \centering
        \includegraphics[width=\textwidth]{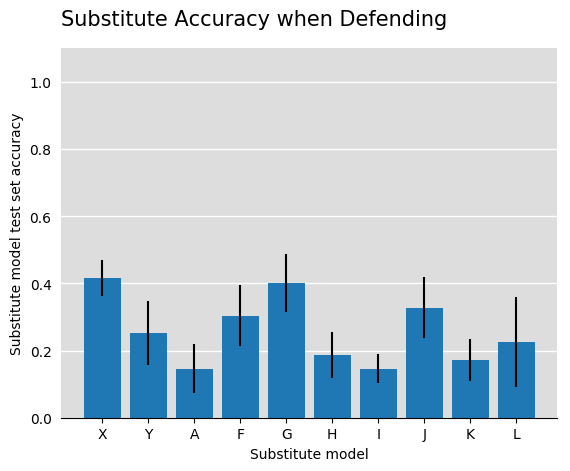}
    \end{subfigure}
    \caption{Results on MNIST. \emph{Top Row:} The black-box classifier's accuracy on adversarial samples generated using the attacker's substitute model. When not defending (left), the classifier suffers severe degradation in performance compared to its baseline accuracy. However, when actively defending (right), the performance loss non-existent or at most minimal across all attack architectures.
    \emph{Bottom Row:} The substitute model's accuracy on test set data. When not defended against (left), the substitute models are able to achieve respectable performance by performing model theft on the black-box model. But when the defender is performing our proposed counter-attack (right), model theft is no longer possible and the substitute models degrade to slightly above random guessing.}
    \label{fig:bars}
\end{figure}

Figure~\ref{fig:bars} displays the results of our experiments.
As can be seen, our proposed counter-attack successfully prevents model theft from all considered substitute model architectures.
Moreover, it corrupts their training procedure so much that they degrade to little more than random guessing, which is impressive, considering that most of them achieve a baseline accuracy of above 60\% simply after training on the provided 150 seed examples.

Further, Table~\ref{tbl:true} shows the fraction of our outputs where the top prediction remains unchanged.
Namely, even when defending against all substitute architectures, over $99.9\%$ of the time our classifier will still return the same label as being most likely than it would have returned if it hadn't defended.
This means that our counter-attack is extremely effective while leaving our non-malicous customers largely unaffected.

\begin{table}[ht!]
    \centering
    \begin{tabular}{crr}
\toprule
ID & Argmax unchanged \\
\midrule
X &    0.999 (0.001) \\
Y &    1.000 (0.000) \\
A &    0.999 (0.001) \\
F &    0.999 (0.001) \\
G &    0.999 (0.001) \\
H &    1.000 (0.001) \\
I &    1.000 (0.000) \\
J &    1.000 (0.000) \\
K &    1.000 (0.000) \\
L &    0.999 (0.000) \\
\bottomrule
\end{tabular}
     \caption{Results on MNIST. Fraction (mean and standard deviation) of outputs where the argmax of the adversarial output distribution is equal to the uncorrupted output distribution. Note that for well over $99.9\%$ of outputs, our top prediction remains unchanged.}
    \label{tbl:true}
\end{table}

\FloatBarrier

\section{Conclusion}
We have shown that black-box attacks based on model theft can successfully be averted by generating adversarial output label distributions that cause a model thief's training gradients to explode.
The generated perturbations do not influence the ordinary use of the black-box model, in particular they do not change the top ranked label in the vast majority of cases.
Future work can extend our method by finding a counter-attack method in the case of binary labels, finding systematic ways for the attacker to avoid being affected by the counter-attack without losing efficiency, as well as defining an adversarial sample generation procedure that incorporates the normalization and positivity constraints on the model outputs.

\bibliography{references}
\bibliographystyle{iclr2018_conference}

\newpage
\appendix
\label{section:app}

\section{Renormalization strategies}
This section presents methods to transform a generated adversarial output $y^*$ into a normalized distribution.
Unfortunately, we have found that naively clipping $y^*$ to $[0, 1]$ and dividing by its sum will cause the sample to cease to be adversarial.
Therefore, we show two methods to perform normalization that we have found to not interfere with the adversarial-ness of $y^*$.

\subsection{Centering the perturbation}
This method alternates between making the adverarial perturbation to be zero mean and clipping the result to the valid range.
We perform the following steps for multiple rounds:
\begin{itemize}
    \item Let $\delta_y = y^* - y$ be the adversarial perturbation of $y^*$
    \item Let $\bar{\delta}_y = \delta_y - \text{mean}(\delta_y)$ be the centered version of $\delta_y$
    \item Let $y^* = y + \bar{\delta}_y$
    \item Clip $y^*$ to $[0, 1]$
\end{itemize}

The number of rounds needed to reach a satisfiably normalized distribution depends on the value of $\epsilon$ used when generating $y^*$ initially.
We have found that for our experiments, between 2 and 5 rounds of the above steps produce $y^*$ with normalization constants off by less than $1/1000$ from $1.0$.

\subsection{Winner takes all}
This method allocates the missing / excess mass to the lowest / highest parts of the distribution.
In the case of missing mass, meaning $\sum_i y^*_i < 1$, we simply allocate the remainder to the lowest entry of $y^*$.
In the case of excess mass, we reduce the highest entry by the amount needed.

While this method will exactly renormalize the distribution, we have found it to weaken our defense in some instances.

\FloatBarrier
\section{Results on CIFAR10}
\begin{figure}[ht!]
    \centering
    \begin{subfigure}[t]{0.49\textwidth}
        \centering
        \includegraphics[width=\textwidth]{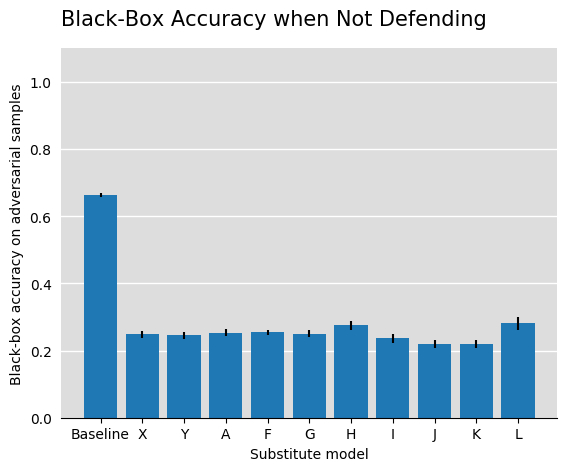}
    \end{subfigure}
    \begin{subfigure}[t]{0.49\textwidth}
        \centering
        \includegraphics[width=\textwidth]{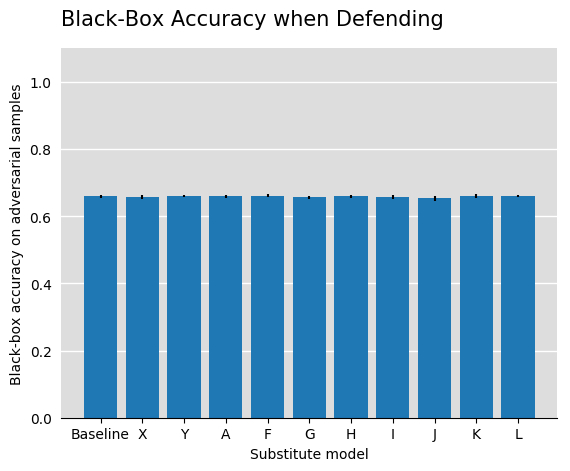}
    \end{subfigure}
    \par\bigskip
    \begin{subfigure}[b]{0.49\textwidth}
        \centering
        \includegraphics[width=\textwidth]{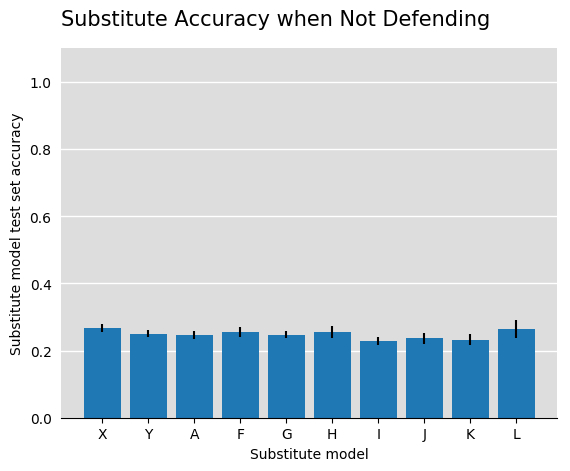}
    \end{subfigure}
    \begin{subfigure}[b]{0.49\textwidth}
        \centering
        \includegraphics[width=\textwidth]{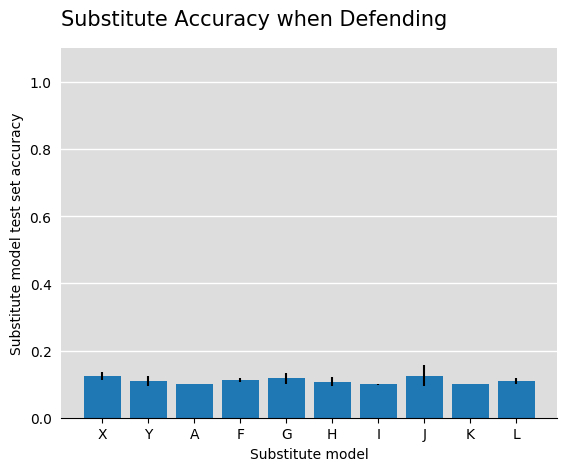}
    \end{subfigure}
    \caption{Results on CIFAR10. \emph{Top Row:} The black-box classifier's accuracy on adversarial samples generated using the attacker's substitute model. 
    \emph{Bottom Row:} The substitute model's accuracy on test set data.}
    \label{fig:bars:cifar10}
\end{figure}

Figure~\ref{fig:bars:cifar10} and Table~\ref{tbl:true:cifar10} show results to experiments on CIFAR10 that are equivalent to the ones on MNIST in the main section.

\begin{table}[ht!]
    \centering
    \begin{tabular}{crr}
\toprule
ID & Argmax unchanged \\
\midrule
X &    1.000 (0.000) \\
Y &    1.000 (0.000) \\
A &    1.000 (0.000) \\
F &    1.000 (0.000) \\
G &    1.000 (0.000) \\
H &    1.000 (0.000) \\
I &    1.000 (0.000) \\
J &    1.000 (0.000) \\
K &    1.000 (0.000) \\
L &    1.000 (0.000) \\
\bottomrule
\end{tabular}
     \caption{Results on CIFAR10. Fraction (mean and standard deviation) of outputs where the argmax of the adversarial output distribution is equal to the uncorrupted output distribution. Note that for well over 99\% of outputs, our top prediction remains unchanged.}
    \label{tbl:true:cifar10}
\end{table}

\FloatBarrier
\section{Results on SVHN}
\begin{figure}[ht!]
    \centering
    \begin{subfigure}[t]{0.49\textwidth}
        \centering
        \includegraphics[width=\textwidth]{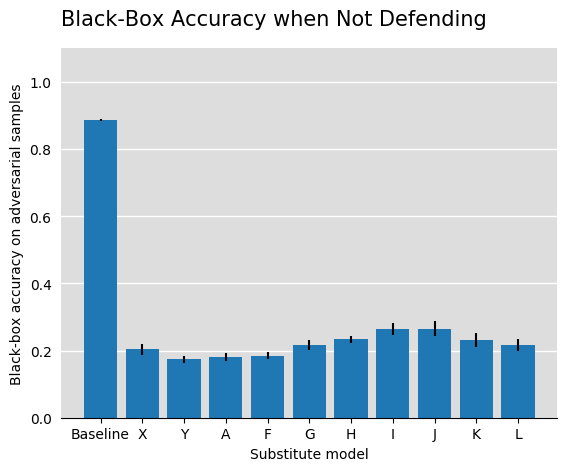}
    \end{subfigure}
    \begin{subfigure}[t]{0.49\textwidth}
        \centering
        \includegraphics[width=\textwidth]{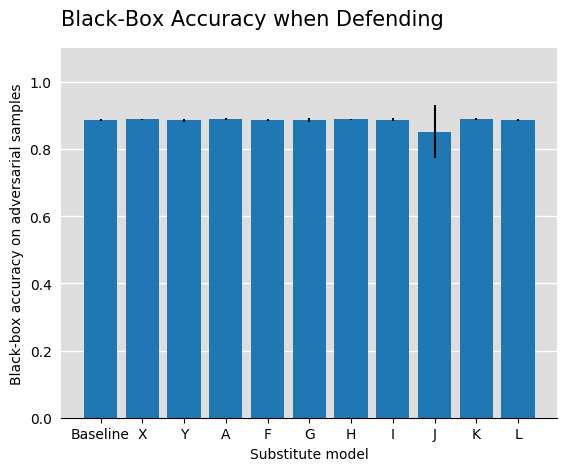}
    \end{subfigure}
    \par\bigskip
    \begin{subfigure}[b]{0.49\textwidth}
        \centering
        \includegraphics[width=\textwidth]{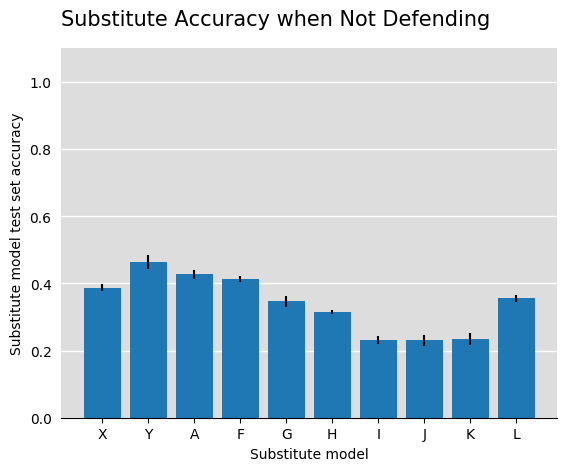}
    \end{subfigure}
    \begin{subfigure}[b]{0.49\textwidth}
        \centering
        \includegraphics[width=\textwidth]{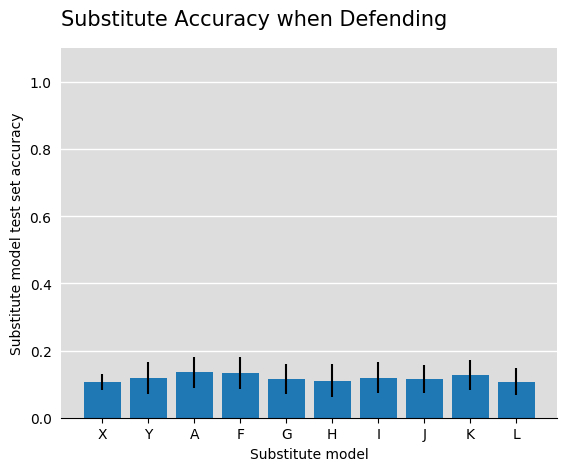}
    \end{subfigure}
    \caption{Results on SVHN. \emph{Top Row:} The black-box classifier's accuracy on adversarial samples generated using the attacker's substitute model. 
    \emph{Bottom Row:} The substitute model's accuracy on test set data.}
    \label{fig:bars:svhn}
\end{figure}

Figure~\ref{fig:bars:svhn} and Table~\ref{tbl:true:svhn} show results to experiments on SVHN that are equivalent to the ones on MNIST in the main section.

\begin{table}[ht!]
    \centering
    \begin{tabular}{crr}
\toprule
ID & Argmax unchanged \\
\midrule
X &    1.000 (0.000) \\
Y &    1.000 (0.000) \\
A &    1.000 (0.000) \\
F &    1.000 (0.000) \\
G &    1.000 (0.000) \\
H &    1.000 (0.000) \\
I &    1.000 (0.000) \\
J &    1.000 (0.000) \\
K &    1.000 (0.001) \\
L &    1.000 (0.000) \\
\bottomrule
\end{tabular}
     \caption{Results on SVHN. Fraction (mean and standard deviation) of outputs where the argmax of the adversarial output distribution is equal to the uncorrupted output distribution. Note that for well over 99\% of outputs, our top prediction remains unchanged.}
    \label{tbl:true:svhn}
\end{table}

\FloatBarrier
 
\end{document}